\documentclass{bmvc2k}

\pdfoutput=1 
\usepackage{multirow}

\usepackage{floatrow}
\newfloatcommand{capbtabbox}{table}[][\FBwidth]

\usepackage{times}

\usepackage{soul}
\usepackage{url}
\usepackage{graphicx}
\usepackage{amsmath}
\usepackage{multirow}
\usepackage{booktabs}
\usepackage{mathrsfs}
\usepackage{color}
\usepackage{subfigure}

\definecolor{OliveGreen}{rgb}{0,0,0}
\newcommand{\tinne}[1]{\textcolor{OliveGreen}{#1}}

\definecolor{airforceblue}{rgb}{0, 0, 0}

\title{ On the Exploration of Incremental Learning for Fine-grained Image Retrieval}

\addauthor{Wei Chen}{w.chen@liacs.leidenuniv.nl}{1}
\addauthor{Yu Liu}{yu.liu@esat.kuleuven.be}{2}
\addauthor{Weiping Wang}{wangwp@nudt.edu.cn}{3}
\addauthor{Tinne Tuytelaars}{Tinne.Tuytelaars@esat.kuleuven.be}{2}
\addauthor{Erwin M. Bakker}{e.m.bakker@liacs.leidenuniv.nl}{1}
\addauthor{Michael Lew}{m.s.k.lew@liacs.leidenuniv.nl}{1}

\addinstitution{
MediaLab, LIACS, \\
 Leiden University,\\
 Leiden, The Netherlands
}
\addinstitution{
 ESAT-PSI,\\
 KU Leuven,\\
 Belgium
 }
 \addinstitution{
 College of Systems Engineering,\\
 NUDT,\\
 Changsha, China
}

\runninghead{CHEN ET AL.}{On the Exploration of Incremental Learning }

\begin{document}

\maketitle

\begin{abstract}

In this paper, we consider the problem of fine-grained image retrieval in an incremental setting, when new categories are added over time. On the one hand, repeatedly training the representation on 
the extended dataset is time-consuming. On the other hand, fine-tuning the learned representation only with the new classes leads to catastrophic forgetting.
To this end, we propose an incremental learning method to mitigate retrieval performance degradation caused by the forgetting issue. Without accessing any samples of the original classes, the classifier of the original network provides soft “labels” to transfer knowledge to train the adaptive network, so as to preserve the previous capability for classification. More importantly, a regularization function based on Maximum Mean Discrepancy is devised to minimize the discrepancy of new classes features from the original network and the adaptive network, respectively. Extensive experiments on two datasets show that our method effectively mitigates the catastrophic forgetting on the original classes while achieving high performance on the new classes.

\end{abstract}

\section{Introduction}
\label{sec:intro}

In an era when the number of images is increasing, deep models for fine-grained image retrieval (FGIR) are required to be adaptable for new incoming classes. However, current image retrieval approaches are focusing mainly on static datasets and are not suited for incremental learning scenarios. To be specific, deep networks well-trained on original classes will under-perform on new incoming classes. 

\begin{figure}[!t]
\centering
 { 
   \includegraphics[width=1\columnwidth]{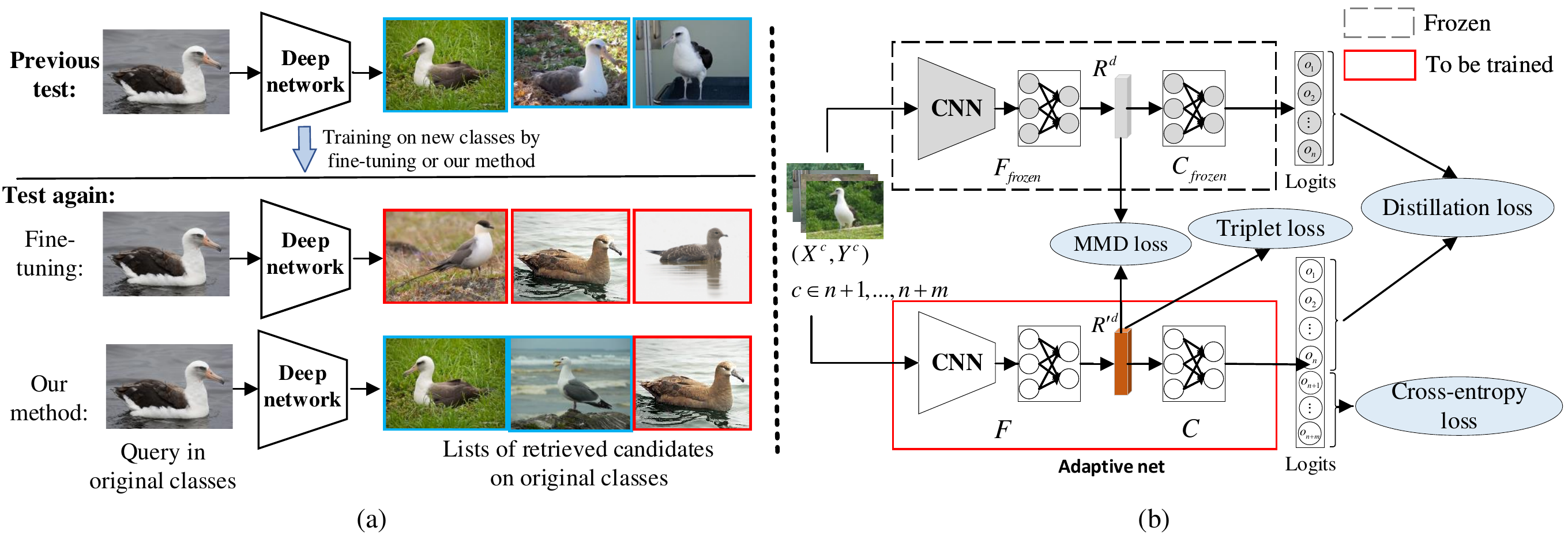} 
 }
 \vspace{-2.3em}
    \caption{ (a) Illustration of catastrophic forgetting for FGIR. Our method aims to maintain good performance on the original classes where the inaccurate returned images are in red box and correct results are in blue box. (b) Framework of our method. The only inputs for the adaptive net $ \mathcal{B} $ are $ m $ new classes and labels $ (\textbf{X}^{c^{\prime}}, \textbf{Y}^{c^{\prime}})$, $ c^{\prime} \!\! \in \!\! (n \!+ \! 1,..., n\! + \! m)$. The frozen net $ \mathcal{A} $ is firstly trained on $ n $ original classes and then copied as initialization for net $ \mathcal{B} $.
}
\label{Whole_Framework} 
\end{figure}

When new classes are added into an existing dataset, joint training on all classes allows to guarantee the performance. However, as the \tinne{number of} new classes increases sequentially, the repetitive re-training is time-consuming. Alternatively, fine-tuning makes the network adapt to new classes and achieve good performance on these classes. However, when the original classes become inaccessible during fine-tuning, the performance of the original classes degrades dramatically because of catastrophic forgetting, a phenomenon that occurs when a network is trained sequentially on a series of new tasks and the learning of these tasks interferes with performance on previous tasks, as shown in Figure \ref{Whole_Framework}(a).

Most of incremental learning methods are exploited for image classification, which is robust and forgiving as long as features remain within the classification boundaries. In contrast, image retrieval focuses more on the discrimination in the feature space rather than the classification decisions. Especially for FGIR, small changes on visual features may have a big impact on the retrieval performance. Additionally, we find that standard methods like Learning without forgetting (\emph{i.e.} LwF \cite{li2017learning}) and Elastic Weight Consolidation (\emph{i.e.} EWC \cite{kirkpatrick2017overcoming}) are insufficient for this problem because the distillation is not on the actual feature space (see Section \ref{One_step_incremental} and \ref{Multi_step_incremental}).

Considering the above limitations, we propose a deep learning model to tackle the problem of incremental fine-grained image retrieval. We regularize the updates of the model to simultaneously retain preservation on original classes and adaptation on new classes. Importantly, to avoid the repeated training, the samples of the original classes are not used when learning the new classes. The classifier of the original network provides soft ``labels'' to transfer knowledge to train the adaptive network using the distillation loss function \cite{hinton2015distilling}\cite{wu2019large}. This focuses on pair-wise similarity but can not well quantify the distance between two feature distributions. This limitation inspires us to adopt a regularization term based on Maximum Mean Discrepancy (MMD) \cite{gretton2012optimal} to minimize the discrepancy between the features derived from an original network and an adaptive network, respectively. Moreover, the cross-entropy loss and triplet loss are utilized to identify subtle differences among sub-categories.

In summary, our contributions are two-fold. First, our work extends FGIR in the context of incremental learning. This is the first work to study this problem, to the best of our knowledge. Second, we propose a deep network, which includes a knowledge distillation loss and a MMD loss, for incremental learning without using any samples from the original classes. It achieves significant improvements over previous incremental learning methods.

\section{Related Work}

Incremental learning is the process of transferring learned knowledge from an original model to an incremental model. It has been researched in a few applications like image classification \cite{li2017learning}\cite{yao2019adversarial}\cite{li2018explicit}\cite{zhou2019m2kd}, image generation \cite{zhai2019lifelong}\cite{xiang2019incremental}, object detection \cite{shmelkov2017incremental}, hashing image retrieval \cite{wu2019deep} and semantic segmentation \cite{michieli2019incremental}. To overcome the so-called catastrophic forgetting, numerous methods have been proposed. For example, a subset of data (exemplars) of original classes are stored into an external memory, and the forgetting is thereby avoided by replaying these exemplars \cite{hou2018lifelong}\cite{lopez2017gradient}\cite{wu2019large}. Recently, GANs \cite{goodfellow2014generative} are used to synthesize samples with respect to the previous data distributions \cite{shin2017continual}\cite{van2018generative}, which avoids the shortcomings of memory-consuming and exemplar-choosing, but generating real-like images with complex semantics is a challenging task. Alternatively, regularization methods constrain the objective functions or parameters of deep networks to preserve the previously learned knowledge. The distillation loss function \cite{hinton2015distilling} is used to transfer knowledge of old classes \cite{li2017learning}\cite{wu2019large}. The importance weight per parameter is estimated based on the old classes, and then is used as regularization to penalize essential parameter changes when training on new incoming classes \cite{kirkpatrick2017overcoming}.

\section{Proposed Approach}

\noindent\textbf{Problem Formulation} Given a fine-grained dataset which includes $ n $ class labels $(\textbf{X}^{c}, \textbf{Y}^{c})$ where $ c \! \in \! (1,..., n)$, each sub-category $ c $ has a different amount of images in $ \textbf{X}^{c} $ and the ground-truth labels $ \textbf{Y}^{c} $. A deep network is trained to perform the retrieval task for the $n$ classes. Consider the incremental learning scenario, images from $ m $ new classes are added sequentially or at once. 
We take as input only the images from $ m $ new incoming classes, \emph{i.e.} $ ( \textbf{X}^{c^{\prime}}, \textbf{Y}^{c^{\prime}}) $, where $c^{\prime} \! \in \! (n \!+ \! 1,..., n\! + \! m)$, to incrementally train the deep network. 
In this way, it is efficient to update the network with no need of re-training the original classes again. Besides, the image instances from the original classes are not always accessible due to privacy issue or memory limit.
Finally, the aim is to continually train the network, to make it preserve promising performance for all seen classes.

\noindent\textbf{Overall Idea} As shown in Figure \ref{Whole_Framework}(b), our method includes two training stages. First, we train a network $ \mathcal{A} $ on the original classes using cross-entropy and triplet loss on the output logits and representations. After $ \mathcal{A} $ is well-trained, we make two copies of $ \mathcal{A} $: one freezing its parameters when incrementally training, and the other adapting its parameters for the $ m $ incremental classes. We refer to this adaptive network as $ \mathcal{B} $. It is initialized with parameters from $ \mathcal{A} $, including the feature extraction layers $ F_{frozen} $ and classifier $ C_{frozen} $, but extends the number of neurons in its classifier $ C $, from which the output logits are $ (o^{\prime}_{1}, o^{\prime}_{2},\dots, o^{\prime}_{n}, o^{\prime}_{n+1},\dots, o^{\prime}_{n+m} ) $, and previous $n$ neurons are copied from $ C_{frozen} $. To overcome catastrophic forgetting, we propose to integrate two regularization strategies based on knowledge distillation and maximum mean discrepancy, respectively.
Given a query image from either the original classes or newly added classes, we extract the features from the fully-connected layer for image retrieval.
We introduce the details of our method below.

\subsection{Semantic Preserving Loss}
First, we train the model with the standard cross-entropy loss. Given the logits $ (o_{1}, o_{2},..., o_{n} ) $ and its class label $ (y_{1}, y_{2},\dots, y_{n} ) $, the loss is $ H(\textbf{y}, \textbf{o}) = -\sum(\textbf{y} \ast log(softmax(\textbf{o})))$. Note that we only use images from the new classes during incremental training, thus the classification is performed on $ (o^{\prime}_{n+1}, o^{\prime}_{n+2},\dots, o^{\prime}_{n+m} ) $, the categorical cross-entropy loss function $ L_{ce} $ is 
\begin{equation}
\label{Modalityclassification_unified}
\begin{aligned}
   L_{ce}    = - \frac{1}{N}\sum^{N}_{i=1}(y_{i} \ast log(\frac{e^{o^{\prime}_{i}(x)}}{\sum^{n+m}_{j=n+1}e^{o^{\prime}_{j}(x)}}))
\end{aligned}
\end{equation}

To identify subtle differences among categories, we adopt the triplet loss $ L_{triplet}$ by mining the hard positive pairs and hard negative pairs based on feature vectors $ \textbf{R} $.

\begin{equation}
\label{Triplet_loss}
\begin{aligned}
   L_{triplet}    = \frac{1}{N}\sum^{N}_{i=1}(\max(0, \lambda + S_{i, neg} - S_{i, pos}))
\end{aligned}
\end{equation}
where $S_{i, neg}$ and $ S_{i, pos}$, based on matrix multiplication (\emph{i.e.} $  S_{i, neg} = R_{i} R^\top_{neg}$), indicate the similarity of $i^{th}$  hard negative and positive pairs, respectively. $ \lambda $ is the margin parameter. 

\subsection{Knowledge Distillation Loss}
We rewrite $( F_{frozen} $, $ C_{frozen} )$ as $( F_{_{f}} $, $ C_{_{f}} )$ for simplicity. Knowledge distillation loss \cite{hinton2015distilling} is defined to regularize the activations of the output layer in both the old and new model. To be specific, we constrain the first $ n $ values in $ (o^{\prime}_{1}, o^{\prime}_{2},..., o^{\prime}_{n}, o^{\prime}_{n+1},..., o^{\prime}_{n+m} ) $ as close as possible to the logits $ (o_{1}, o_{2},..., o_{n}) $ from the frozen network $ \mathcal{A} $. Following the method in \cite{wu2019large}\cite{li2017learning}, when $ m $ new classes are added at once, we compute the knowledge distillation loss by

\begin{equation}
\label{Distillation_loss}
\begin{aligned}
L_{dist} \! &   = \displaystyle - \frac{1}{|\textbf{X}^{c^{\prime}}|} \sum_{x \in \textbf{X}^{c^{\prime}}}^{|\textbf{X}^{c^{\prime}}|} \sum_{k=1}^{n}  \{ p_{k}(x) \ast log[p^{\prime}_{k}(x)] \} \\
\end{aligned}
\end{equation}
where $ p_{k}(x) \! = \! \frac{e^{o_{k}(x)/T}}{\sum^{n}_{j}e^{o_{j}(x)/T}}$ and $ p^{\prime}_{k}(x) \! = \! \frac{e^{o^{\prime}_{k}(x)/T}}{\sum^{n}_{j}e^{o^{\prime}_{j}(x)/T}} $, $ T $ is a temperature factor that is normally set to 2 \cite{li2017learning}. $ \textbf{p} \! = \! \{p\}_{n} $ and $ \textbf{p}^{\prime} \! = \! \{p^{\prime} \}_{n} $ refer to the probabilities produced by the modified Softmax function in \cite{hinton2015distilling}. $F_{_{f}} $ and $ C_{_{f}} $ denote the parameters of network $ \mathcal{A} $. Similarly, $ F $ and $ C $ denote the parameters of network $ \mathcal{B} $, as shown in Figure \ref{Whole_Framework}(b). $| \textbf{X}^{c^{\prime}} |$ indicates the number of images from the new $ m $ classes in a mini-batch. $ n $ denotes the number of the original classes. Note that $n$ will be extended accordingly when more new classes are added. 

\begin{figure}[!t]
\centering  
  \subfigure[] 
 { \label{MMD_explanation_theory}   
  \includegraphics[width=0.4\columnwidth]{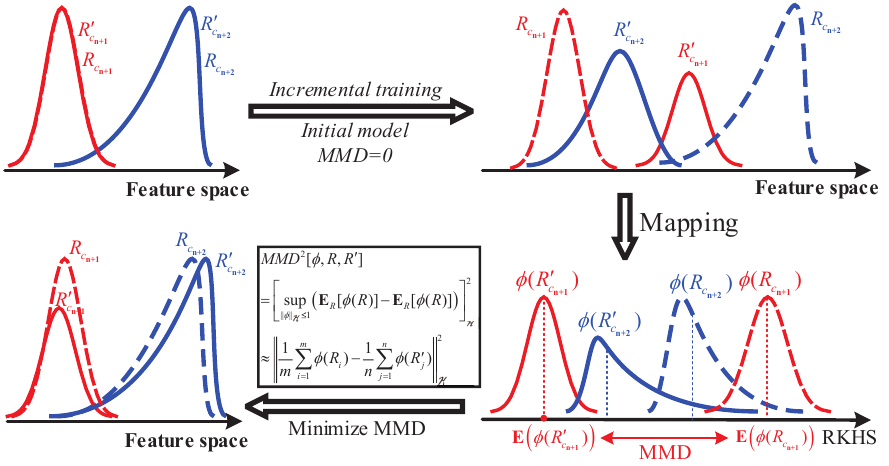}    
  }  
    \subfigure[]
 { \label{MMD_explanation}     
   \includegraphics[width=0.5\columnwidth]{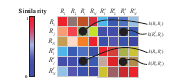}
 }  
    \vspace{-1.5em}
    \caption{(a)  The red and blue color depict the feature distributions of two categories. The dashed line indicates the distributions from the network $ \mathcal{A} $, the solid line indicates that from the network $ \mathcal{B} $. Since $ \mathcal{A} $ is copied as the initial model for network $ \mathcal{B} $, MMD=0 in the beginning. As training progresses, $ \mathcal{B} $ is updated to change its output features and the MMD is expected to increase. (b) MMD for instance-to-instance similarity. } 

 \end{figure}

\subsection{ Maximum Mean Discrepancy Loss}

Knowledge distillation loss focuses on constraining classification boundaries to  mitigate the forgetting issue. However, for FGIR, it is more important to reduce the difference between feature distributions. For this, we adopt maximum mean discrepancy (MMD) \cite{gretton2012optimal} to capture the correlation of feature distributions between network $ \mathcal{A} $ and $ \mathcal{B} $. MMD has been used to bridge source and target distributions such as in domain adaptation \cite{long2016unsupervised}\cite{yan2017mind}. However, our work is the first to impose MMD to regularize the forgetting issue for FGIR. 

Given the features $ R^{d} $ (d is feature dimension) from network $ \mathcal{A} $ and $ \mathcal{B} $, MMD measures the distance between the means of two feature distributions after mapping them into a reproducing kernel Hilbert space (RKHS). In Figure \ref{MMD_explanation_theory}, we illustrate how MMD mitigates the catastrophic forgetting issue. Note that, in the Hilbert space $ \mathcal{H} $, norm operation can be equal to the inner product \cite{chwialkowski2016kernel}\cite{gretton2012optimal}. Finally, the squared MMD distance is:
\begin{equation}
\label{Modalityclassification_expension}
\begin{aligned}
& \rm{MMD}^{2}(\textbf{R}, \textbf{R}^{\prime}) = || \frac{1}{N} \sum^{N}_{i=1} \! \phi(R_{i}) \! - \! \frac{1}{N} \sum^{N}_{j=1}\! \phi(R^{\prime}_{j})||^{2}_{\mathcal{H}} \\
& = \frac{1}{N^{2}}  \!\! < \!\! \sum^{N}_{i=1} \! \phi(R_{i}) \! - \!  \sum^{N}_{j=1}\! \phi(R^{\prime}_{j}),  \sum^{N}_{i=1} \! \phi(R_{i}) \! - \!  \sum^{N}_{j=1}\! \phi(R^{\prime}_{j}) \!\!>_{_{\mathcal{H}}}  \\
& = \! \frac{1}{N^{2}} \Big [ \! \sum^{N}_{i=1} \! \sum^{N}_{j=1} \!\!\!\! < \!\! \phi(R_{i}), \phi(R_{j}) \!\! >_{_{\mathcal{H}}} \!\!  + \! \sum^{N}_{i=1} \! \sum^{N}_{j=1} \!\! < \!\! \phi(R^{\prime}_{i}), \phi(R^{\prime}_{j}) \!\! >_{_{\mathcal{H}}} \!\! - 2 \sum^{N}_{i=1} \! \sum^{N}_{j=1} \!\!\!\! < \!\! \phi(R_{i}), \phi(R^{\prime}_{j}) \!\! >_{_{\mathcal{H}}}   \!\!  \Big ] \\
& \quad \quad s.t. \ R = F_{_{f}}(x), \ R^{\prime}=F(x)
\end{aligned}
\end{equation}
where  $ N $ is batch size, and $ \phi( \cdot) $ denotes the mapping function. However, it is hard to determine $ \phi( \cdot) $. In RKHS, the kernel trick is used to replace the inner product in Eq. \ref{Modalityclassification_expension}, \emph{i.e.} $ \!\! < \!\! \phi(R),\phi(R^{\prime}) \!\! >  = \!  k(R, R^{\prime})  $. Considering all the features in a mini-batch, $ \textbf{R} $ \!\!\! = \!\!\! \{$ R \}_{N}$ and $ \textbf{R}^{\prime} $ \!\!\! = \!\!\! \{$ R^{\prime}   \}_{N} $, we define the MMD loss $ L_{mmd} $ as:
\begin{equation}
\label{Modalityclassification_unified}
\begin{aligned}
 L_{mmd} =  \rm{MMD}(\textbf{R}, \textbf{R}^{\prime}) =   
 \displaystyle \frac{1}{N} \Big [  \sum^{N}_{i=1} \! \sum^{N}_{j=1} \! k(R_{i}, R_{j}) \!-  2 \!\sum^{N}_{i=1} \! \sum^{N}_{j=1} \!\! k(R_{i}, R^{\prime}_{j}) \! + \!\!\sum^{N}_{i=1} \! \sum^{N}_{j=1} \!\! k(R^{\prime}_{i}, R^{\prime}_{j}) \Big ]^{\frac{1}{2}} \\
\end{aligned}
\end{equation}
where $ \rm{k}(R,R^{\prime}) \!\! = \!\! \rm{exp}(-(|| R \! - \! R^{\prime} ||^{2}_{2}) / (2 \sigma^{2}_{m})) $, $ \sigma_{m} $ means $ m $ variances in the Gaussian kernel. 

\noindent\textbf{Discussion}. Knowledge distillation loss focuses on constraining pair-wise similarity. However, MMD loss measures the distance between each feature vector, as depicted in Figure \ref{MMD_explanation}. Finally, it captures the distance of two feature distributions from the frozen net and adaptive net. Thus, MMD loss is more powerful to quantize the correlation of two models.
 
Overall, the objective function in our method for incremental FGIR learning is:
\begin{equation}
\label{Over_all_loss_functions}
\begin{aligned}
   L   = \alpha L_{dist} + \beta L_{mmd} +  (L_{ce} + L_{triplet})
\end{aligned}
\end{equation}

\section{Experiments}
\subsection{Datasets and Experimental Settings}
\textbf{Datasets}. We demonstrate our method on the Stanford-Dogs \cite{khosla2011novel} and CUB-Birds \cite{wah2011caltech} datasets. For the former, we use the official train/test splits. When training incrementally, we split the first 60 sub-categories (in the order of official classes) as the original classes and images from the remaining 60 sub-categories are added at once or sequentially. For the latter, we choose 60\% of images from each sub-category as training set and 40\% as testing set. Afterwards, we split the first 100 sub-categories (in the order of official classes) as the original classes and the remaining 100 sub-categories as new classes. The details are shown in Table 1. 
\begin{table}[h] 
\label{Statistics_of_datasets}
\centering
\scriptsize
\setlength{\tabcolsep}{0.7mm}
\begin{tabular}{|l|l|c|c|c|c|c|c|}
\hline
\multicolumn{2}{|c|}{\multirow{2}{*}{Datasets}} & \multicolumn{3}{c|}{\begin{tabular}[c]{@{}c@{}}Training set\\ (\#Image/\#Class)\end{tabular}} & \multicolumn{3}{c|}{\begin{tabular}[c]{@{}c@{}}Testing set\\ (\#Image/\#Class)\end{tabular}} \\ \cline{3-8} 
\multicolumn{2}{|c|}{}   & Original cls.  & New cls.  & Total  & Original cls.  & New cls.   & Total    \\
 \hline
\multicolumn{2}{|l|}{Stanford-Dogs}  & 6000/60  &  6000/60  &  12000/120 &  4651/60   &  3929/60   &  8580/120    \\ 
\hline
\multicolumn{2}{|l|}{CUB-Birds}    & \multicolumn{1}{c|}{3504/100}   & \multicolumn{1}{c|}{3544/100}  & \multicolumn{1}{c|}{7048/200}  & \multicolumn{1}{c|}{2360/100} & \multicolumn{1}{c|}{2380/100}  & \multicolumn{1}{c|}{4740/200}  \\ 
\hline
\end{tabular}
\vspace{-2em}
\caption{ Statistics of the datasets used in our experiments.}
\end{table}

$ \!\!\!\!\!\!\!\!\!\!\!\!\!\!\!\!\! $ \textbf{Experimental Settings}. We use the Recall@K \cite{jegou2010product}\cite{oh2016deep} (K is the number of retrieved samples),  mean Average Precision (mAP), the precision-recall (PR) curve and feature distribution visualizations for evaluation. We adopt the Google Inception \cite{szegedy2015going} to extract image features. During training, the parameters in Inception are updated using the Adam optimizer \cite{kingma2014adam} with a learning rate of $1 \times 10^{-6} $, while parameters in fully-connected layers and classifier are updated with a learning rate of $1 \times 10^{-5} $. We follow the sampling strategy in \cite{wang2019multi} and each incremental process is trained 800 epochs. Following the practice in \cite{oh2016deep}\cite{wang2019multi}, the output 512-D features $ (R^{d})$ from fully-connected layers are used for retrieval. We set the hyper-parameters $ \alpha  \! = \! \beta  \! = \! 1 $ in Eq. \ref{Over_all_loss_functions}, and the margin $ \lambda = 0.5$ in Eq. \ref{Triplet_loss}. Note that we mainly report the results tested on the CUB-Birds dataset in main paper. For the results of the Stanford-Dogs dataset, we show them in the \textit{supplementary material}.

\subsection{One-step Incremental Learning for FGIR}
\label{One_step_incremental}
We report the results for multiple classes added at once. The process includes two stages. First, we use the cross-entropy and triplet loss to train the network $ \mathcal{A} $ on the original classes (100 classes for the CUB-Birds dataset), denoted as $ \mathcal{A}(1 \verb|-| 100)$. Second, only images of new classes are added at once to train network $ \mathcal{B}$, denoted as  $ \mathcal{B}(101 \verb|-| 200)$. DIHN \cite{wu2019deep} has been explored the incremental learning for hashing-based image retrieval. However, its main difference with ours is to depend on the usage of old data as query set to avoid forgetting in their assumption. Considering no previous works for the fine-grained incremental image retrieval, we apply Learning without Forgetting (LwF) \cite{li2017learning}, Elastic Weight Consolidation (EWC) \cite{kirkpatrick2017overcoming}, ALASSO \cite{park2019continual}, and the incremental learning for semantic segmentation (dubbed L2 loss) \cite{michieli2019incremental} for comparison. LwF, EWC, and ALASSO distill knowledge on classifier and network parameters, which are insufficient for incremental FGIR. L2 loss in \cite{michieli2019incremental} is more similar with ours where the knowledge is distilled on the classifier and intermediate feature space. Note that cross-entropy and triplet loss (\emph{i.e.} semantic preserving loss) are combined with these three algorithms for fair comparison. The Recall@K are reported in Table \ref{Results_Table1}.

\begin{table}[!h] 
\centering
\scriptsize
\setlength{\tabcolsep}{0.9mm}
\begin{tabular}{|p{110 pt}|c|c|c|c|c|c|c|}
\hline
\multicolumn{1}{|c|}{Configurations} & \multicolumn{3}{c|}{Original classes} & \multicolumn{3}{c|}{New classes} \\ \cline{1-7} 
\multicolumn{1}{|c|}{Recall@K(\%)}  & K=1  & K=2 & K=4 & K=1 & K=2 & K=4 \\ \hline
$\mathcal{A}$(1-100) (initial model)  &   79.41 & 85.64 & 89.63 & - & - & -  \\
\hline 
+$\mathcal{B}$(101-200) w feature extraction   & - & - & - & 47.02 & 57.44 & 67.86 \\
+$\mathcal{B}$(101-200) w fine-tuning   & 53.90 & 64.56 & 73.56 & 76.18 & 82.56 & 87.39 \\ 
+$\mathcal{B}$(101-200) w LwF ($ L_{dist} $)   & 54.92 & 66.40 & 75.42 & 75.76 & 82.69 & 86.93 \\
+$\mathcal{B}$(101-200) w ALASSO    & 56.91 & 66.65 & 76.57 & 72.48 & 79.50 & 85.67 \\ 
+$\mathcal{B}$(101-200) w EWC    & 62.03 & 72.16 & 80.08 & 73.32 & 80.92 & 86.01 \\ 
+$\mathcal{B}$(101-200) w L2 loss   & 66.48 & 75.68 & 82.67 & \textbf{77.44} & \textbf{83.78} & \textbf{88.07} \\ 
+$\mathcal{B}$(101-200) w  Our method  & \textbf{74.41} & \textbf{82.57 }& \textbf{88.52} & 73.11 & 80.84 & 86.64 \\ 
\hline         
$\mathcal{A}$(1-200) (reference model) & 77.33 & 85.08 & 89.03 & 76.64 & 83.53 & 89.12 \\     
\hline        
\end{tabular}
\vspace{-2em}
\caption{ Recall@K (\%) of incremental FGIR on the CUB-Birds dataset when new classes are added at once. The best performance in the original class and the new class are in bold. }
\label{Results_Table1} 
\end{table}

The ``w feature extraction'' depicts when $\mathcal{A} $ directly extracts features on the new classes without re-training. The ``w fine-tuning'' depicts using $ L_{ce}$ and $ L_{triplet}$ to train $\mathcal{A} $ on the new classes but without using $ L_{dist}$. Overall, the network $ \mathcal{B} $ suffers from the catastrophic forgetting issue and has lower performance on the original classes, whereas our method outperforms the others. As for the new classes, other three algorithms outperform ours. For example, `` w L2 loss'' method achieves on Recall@1 by 4.33\% compared to ours (77.44\% $ \!\! \to \!\! $ 73.11\%). However, it suffers from significant performance degradation on the original classes with Recall@1 dropping by 12.93\% compared to the initial model (79.41\% $ \!\! \to \!\! $ 66.48\%). For our method, the Recall@1 on the original classes is 74.41\% (dropped by 5.00\% from 79.41\% of  the initial model); the Recall@1 on the new classes is 73.11\% compared to the reference model from $ \mathcal{A}(1\verb|-|200)$ (\emph{i.e.} Recall@1=76.64\%). 

\begin{figure}[!b]
\centering  
  \subfigure[]
 { \label{PR_original_cub}     
   \includegraphics[width=0.215\columnwidth]{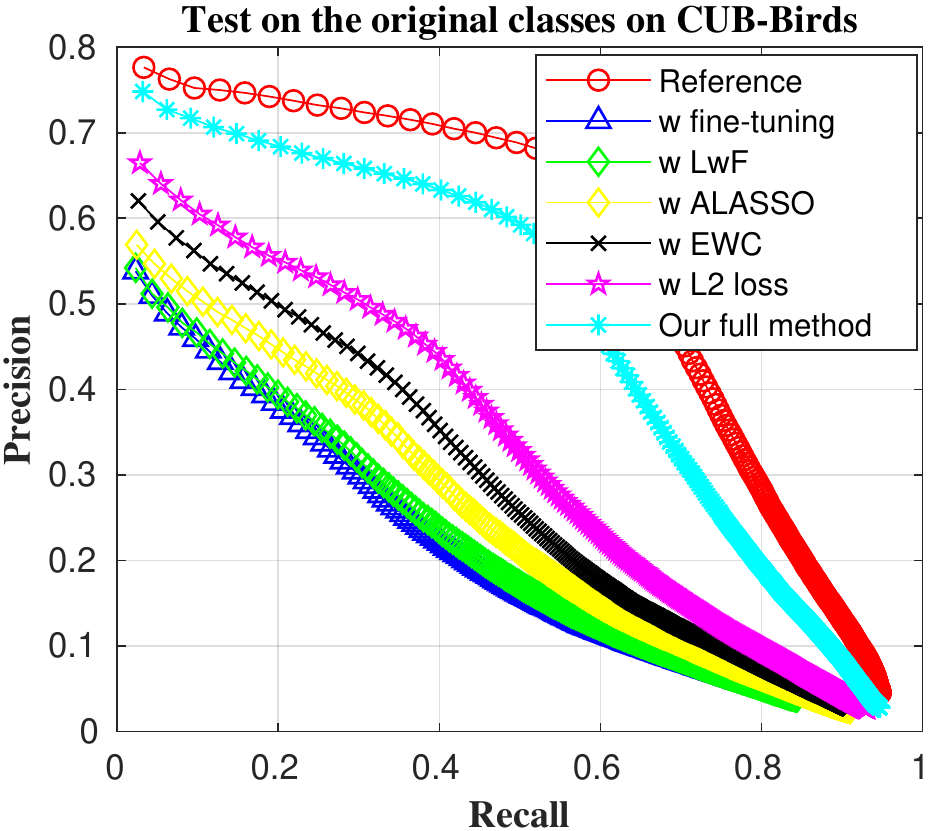}  
 }     
  \subfigure[] 
 { \label{PR_new_cub}   
   \includegraphics[width=0.215\columnwidth]{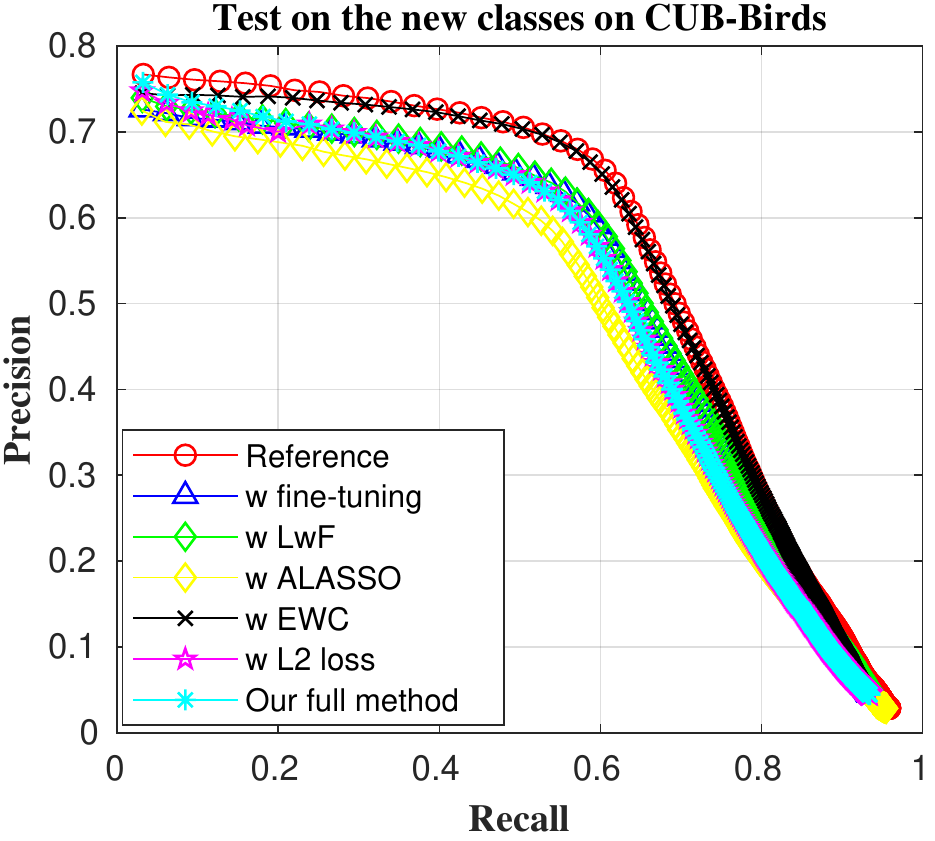}   
  }  
  \subfigure[] 
 { \label{mAP_original_cub}     
   \includegraphics[width=0.215\columnwidth]{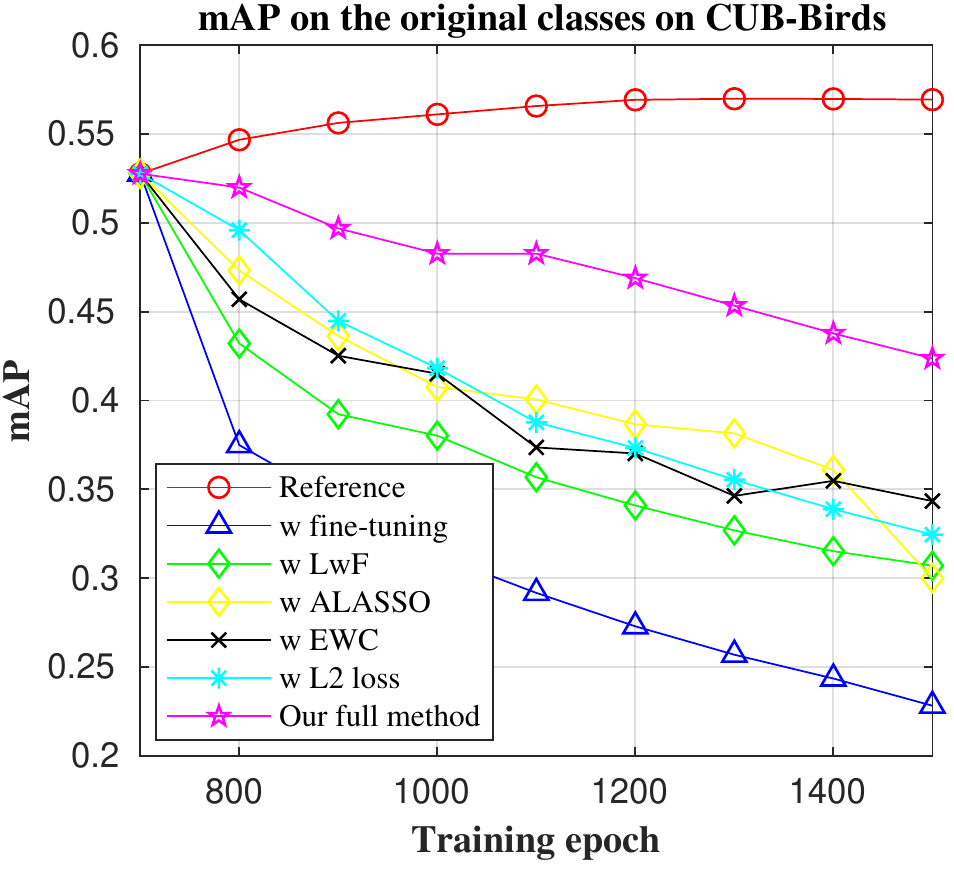}
 }  
  \subfigure[] 
 { \label{training_time_cub}     
   \includegraphics[width=0.235\columnwidth]{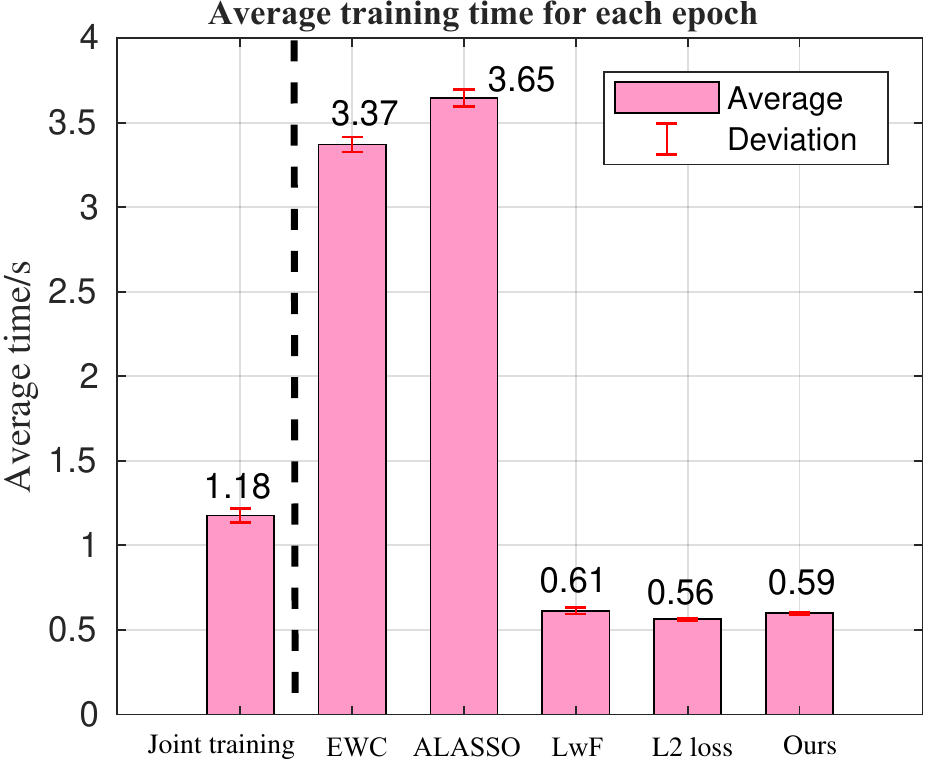} 
 }  
    \vspace{-1.5em}
    \caption{ (a)-(b) denote the PR curves tested on the original classes and new classes. (c) depicts the mAP results for different methods as the training proceeds. We only show the results tested on the original classes. (d) training time comparison during each epoch. }  
\label{Precision_curve_all}  
 \end{figure}

We report the PR curves and mAP results in Figure \ref{PR_original_cub}, \ref{PR_new_cub}, and \ref{mAP_original_cub}, respectively. Overall, when tested on the new classes, all methods share similar trends. When tested on the original classes, our method has better performance although it still has gap to reference performance. For mAP results, the reference results are the same as in Table \ref{Results_Table1}. We utilize the well-trained network $ \mathcal{A}$ at epoch=700 as initial model to train $ \mathcal{B}$ on the new classes until convergence, we test the mAP of network $ \mathcal{B}$ on the original classes. As the curves show, the network trends to degrade its accuracy on the original classes during incremental training.

Furthermore, we explore the influence of the new classes number. Specifically, we choose 100 classes and 25 classes as new categories. The results are reported in Table \ref{Adding_number_cub_p1}, we observe that larger newly-added classes lead to heavier forgetting. For example, when only 25 new classes are used, the Recall@1 drops from 79.41\% to 76.65\%, compared to the one drops from 79.41\% to 74.41\% where 100 new classes are added. Note that the reference models are trained jointly on all classes and tested on the original and new classes separately.

\subsection{Multi-step Incremental Learning for FGIR}
\label{Multi_step_incremental}
We split the new classes into 4 groups and added each group sequentially. The training procedures are as follows: the initial model $\mathcal{A}$ is pre-trained on the original classes (1-100), and used as an initial model to train on newly-added classes (101-125) until convergence to produce a new model $\mathcal{B}$(101-125). Afterwards, the newly-trained model $\mathcal{B}$(101-125) is used as an initial model to train on other new classes (126-150) to produce $\mathcal{B}$(101-125)(126-150). This process is repeated until 4 groups of classes are added sequentially.

We compare to three representative methods (we choose EWC rather than ALASSO since EWC obtains higher performance on the CUB-Birds dataset) and report the results in Table \ref{Results_Table3}. The reference performances are achieved by jointly training all the classes, and then tested on each group (including the original classes). Overall, the model suffers from the catastrophic forgetting issue  when sequentially training. However, our method achieves a minimal performance degradation. For instance, when 4 groups have been added, the model $\mathcal{B}$(101-125)(126-150)(151-175)(176-200) is tested on the original classes(1-100). The ``L2 loss'' algorithm Recall@1 drops 79.41\% $\!\! \to \!\!$ 67.37\% $\!\! \to \!\! $ 58.14\% $\!\! \to \!\!$ 53.86\% $\! \to \!$ 45.85\%, the average degradation is 8.39\%. Our method Recall@1 drops 79.41\% $\!\! \to \!\!$ 76.65\% $\!\! \to \!\!$ 73.77\% $\!\! \to \!\!$ 70.47\% $\!\! \to \!\!$ 66.40\%. The average performance degrades by 3.25\%, which indicates that our method significantly mitigates the forgetting problem. Furthermore, our method has good performance on new classes, which are closer to the reference performance. When the model $\mathcal{B}$(101-125)(126-150)(151-175)(176-200) is tested on new classes (176-200), the results are achieved with Recall@1=85.21\%, Recall@2=89.92\% and Recall@4=93.28\%, respectively, while the reference results are Recall@1=83.70\%, Recall@2=90.25\% and Recall@4=93.78\%.

\begin{table*}[!t]
\tiny
\centering
\setlength{\tabcolsep}{0.75mm}
\begin{tabular}{|p{28pt}|l|c|c|c|c|c|c|c|c|c|c|c|c|c|c|c|}
\hline
\multicolumn{2}{|c|}{Configurations} & \multicolumn{3}{c|}{Original (1-100)} & \multicolumn{3}{c|}{Added new (101-125)} & \multicolumn{3}{c|}{Added new (126-150)} & \multicolumn{3}{c|}{Added new (151-175)} & \multicolumn{3}{c|}{Added new (176-200)}\\ \cline{1-17} 

\multicolumn{2}{|c|}{Recall@K(\%)} & \multicolumn{1}{c|}{K=1} & \multicolumn{1}{c|}{K=2} & \multicolumn{1}{c|}{K=4} & \multicolumn{1}{c|}{K=1} & \multicolumn{1}{c|}{K=2} & \multicolumn{1}{c|}{K=4}& \multicolumn{1}{c|}{K=1} & \multicolumn{1}{c|}{K=2} & \multicolumn{1}{c|}{K=4} & \multicolumn{1}{c|}{K=1} & \multicolumn{1}{c|}{K=2} & \multicolumn{1}{c|}{K=4} & \multicolumn{1}{c|}{K=1} & \multicolumn{1}{c|}{K=2} & \multicolumn{1}{c|}{K=4} \\ \hline

\multicolumn{2}{|c|}{$\mathcal{A}$(1-100) (initial model)} & \multicolumn{1}{c|}{79.41} & \multicolumn{1}{c|}{85.64} & \multicolumn{1}{c|}{89.63}& \multicolumn{1}{c|}{-} & \multicolumn{1}{c|}{-} & \multicolumn{1}{c|}{-}  & \multicolumn{1}{c|}{-}  & \multicolumn{1}{c|}{-}  & \multicolumn{1}{c|}{-} & \multicolumn{1}{c|}{-}  & \multicolumn{1}{c|}{-}  & \multicolumn{1}{c|}{-}  & \multicolumn{1}{c|}{-}  & \multicolumn{1}{c|}{-}  & \multicolumn{1}{c|}{-}       \\ \hline

\multirow{5}{*}{\begin{tabular}[c]{@{}c@{}}LwF\\ algorithm \cite{li2017learning} \end{tabular}} & \multicolumn{1}{l|}{+$\mathcal{B}$(101-125)} & \multicolumn{1}{c|}{57.50} & \multicolumn{1}{c|}{68.05} & \multicolumn{1}{c|}{75.68}    & \multicolumn{1}{c|}{79.59} & \multicolumn{1}{c|}{85.88} & \multicolumn{1}{c|}{88.95} & \multicolumn{1}{c|}{-} & \multicolumn{1}{c|}{-} & \multicolumn{1}{c|}{-} & \multicolumn{1}{c|}{-} & \multicolumn{1}{c|}{-} & \multicolumn{1}{c|}{-} & \multicolumn{1}{c|}{-} & \multicolumn{1}{c|}{-}  & \multicolumn{1}{c|}{-}\\

 &  +$\mathcal{B}$(101-125)(126-150) &   42.46  & 54.03 & 64.66 & 62.59 & 74.83 & 82.31 & 70.17 & 79.67 & 86.00  & - & - & - & - & - & -  \\
 &  +$\mathcal{B}$(101-125)(126-150)(151-175) &  40.21 & 51.57 & 61.27 & 47.79 & 63.10 & 75.68 & 56.83 & 67.33 & 78.17  & 81.57 & 87.27 & 90.79 & - & - & -  \\
 &  +$\mathcal{B}$(101-125)(126-150)(151-175)(176-200) &  33.31  & 44.75 & 55.38 & 49.83 & 63.78 & 75.85 & 48.00 & 60.33 & 72.33  & 67.17 & 75.88 & 82.91 & 83.70 & 89.41 & 92.94  \\
 \cline{2-17} 
  \hline 
  
\multirow{5}{*}{\begin{tabular}[c]{@{}c@{}}EWC\\  algorithm \cite{kirkpatrick2017overcoming} \end{tabular}} & \multicolumn{1}{l|}{+$\mathcal{B}$(101-125)} & \multicolumn{1}{c|}{61.23} & \multicolumn{1}{c|}{70.85} & \multicolumn{1}{c|}{80.04}    & \multicolumn{1}{c|}{80.95} & \multicolumn{1}{c|}{86.39} & \multicolumn{1}{c|}{90.82} & \multicolumn{1}{c|}{-} & \multicolumn{1}{c|}{-} & \multicolumn{1}{c|}{-} & \multicolumn{1}{c|}{-} & \multicolumn{1}{c|}{-} & \multicolumn{1}{c|}{-} & \multicolumn{1}{c|}{-} & \multicolumn{1}{c|}{-}  & \multicolumn{1}{c|}{-}\\

 &  +$\mathcal{B}$(101-125)(126-150) &   46.65  & 56.40 & 67.54 & 65.48 & 77.72 & 84.01 & 72.33 & 80.67 & 86.67  & - & - & - & - & - & -  \\
 &  +$\mathcal{B}$(101-125)(126-150)(151-175) &   43.60 & 54.79 & 64.70 & 61.50 & 72.45 & 80.44 & 66.50 & 75.50 & 82.67  & 81.08 & 85.26 & 87.77 & - & - & -  \\
 &  +$\mathcal{B}$(101-125)(126-150)(151-175)(176-200) &   36.82 & 47.54 & 59.66 & 57.99 & 67.01 & 76.87 & 50.67 & 64.67 & 77.67  & 64.15 & 74.87 & 81.24 & 82.02 & 86.39 & 90.42  \\
 \cline{2-17} 
  \hline 
  
\multirow{5}{*}{\begin{tabular}[c]{@{}c@{}}L2 loss\\  algorithm \cite{michieli2019incremental} \end{tabular}} & \multicolumn{1}{l|}{+$\mathcal{B}$(101-125)} & \multicolumn{1}{c|}{67.37} & \multicolumn{1}{c|}{76.27} & \multicolumn{1}{c|}{83.31}    & \multicolumn{1}{c|}{80.61} & \multicolumn{1}{c|}{85.54} & \multicolumn{1}{c|}{89.46} & \multicolumn{1}{c|}{-} & \multicolumn{1}{c|}{-} & \multicolumn{1}{c|}{-} & \multicolumn{1}{c|}{-} & \multicolumn{1}{c|}{-} & \multicolumn{1}{c|}{-} & \multicolumn{1}{c|}{-} & \multicolumn{1}{c|}{-}  & \multicolumn{1}{c|}{-}\\

 &  +$\mathcal{B}$(101-125)(126-150) &  58.14  & 68.31 & 76.78 & 72.11 & 80.44 & 87.41 & 73.33 & 82.17 & 88.67  & - & - & - & - & - & -  \\
 &  +$\mathcal{B}$(101-125)(126-150)(151-175) &   53.86 & 62.03 & 71.91 & 60.37 & 71.43 & 80.27 & 66.33 & 76.67 & 84.67  & 81.24 & 87.27 & 90.95 & - & - & -  \\
 &  +$\mathcal{B}$(101-125)(126-150)(151-175)(176-200) &   45.85 & 56.61 & 67.75 & 57.65 & 71.77 & 80.95 & 59.33 & 70.50 & 79.13  & 73.70 & 83.08 & 88.94 & 84.20 & 89.24 & 92.10  \\
 \cline{2-17} 
  \hline 
  
\multirow{4}{*}{\begin{tabular}[c]{@{}c@{}}Our method \end{tabular}} & \multicolumn{1}{l|}{+$\mathcal{B}$(101-125)} & \multicolumn{1}{c|}{76.65} & \multicolumn{1}{c|}{83.47} & \multicolumn{1}{c|}{88.86} & \multicolumn{1}{c|}{73.13} & \multicolumn{1}{c|}{82.31} & \multicolumn{1}{c|}{88.44} & \multicolumn{1}{c|}{-} & \multicolumn{1}{c|}{-} & \multicolumn{1}{c|}{-} & \multicolumn{1}{c|}{-} & \multicolumn{1}{c|}{-} & \multicolumn{1}{c|}{-} & \multicolumn{1}{c|}{-} & \multicolumn{1}{c|}{-} & \multicolumn{1}{c|}{-}\\

 &  +$\mathcal{B}$(101-125)(126-150) &   73.77  & 81.36 & 87.80 & 74.32 & 83.33 & 89.29 & 74.50 & 83.00 & 87.83  & - & - & - & - & - & -  \\
 &  +$\mathcal{B}$(101-125)(126-150)(151-175) &   70.47 & 78.77 & 85.97 & 70.41 & 80.78 & 88.78 & 72.00 & 79.17 & 86.83  & 78.89 & 86.77 & 90.26 & - & - & -  \\
 &  +$\mathcal{B}$(101-125)(126-150)(151-175)(176-200) &   66.40 & 75.93 & 83.14 & 70.07 & 80.27 & 86.22 & 69.00 & 78.33 & 85.50 & 73.87 & 83.92 & 88.78 & 85.21 & 89.92 & 93.28  \\
 \cline{2-17} 
  \hline 
  
\multicolumn{2}{|c|}{$\mathcal{A}$(1-200) (reference model)}  &  77.33  & 85.08  &  89.03 & 76.87  & 84.86  & 90.48 & 73.00  &   80.00 & 87.67  &  83.25 & 88.94  & 92.29 &  83.70 & 90.25  &  93.78                     
\\ \hline
\end{tabular}
    \vspace{-2em}
    \caption{ Recall@K (\%) results on the CUB-Birds dataset when new classes are added sequentially. ``Added new (101-125)'' indicates the first 25 classes (101-125) are used as the first part to train the network.}
\label{Results_Table3}
\end{table*}

\subsection{Validation with Image Classification}

We evaluate the effectiveness of our method on the CIFAR-100 dataset \cite{krizhevsky2009learning} which is the popular benchmark for class-incremental learning in image classification. We split 100 classes into a sequence of 5 tasks, and each task includes 20 classes. In Table~\ref{results_for_classification}, the results indicate the average top-1 accuracy of the classes from seen tasks. In the last column, the test set evaluates the classes from all the five tasks. Note that, the 20 classes in the first task (the second column) achieve the same performance, as it has no incremental learning yet. We observe that our method outperforms other methods across the tasks. It suggests our method generalizes well to various applications. Notably, our improvement for image retrieval is more significant than that for image classification. The reason is that the proposed MMD loss is imposed on the feature representation, which largely benefits the metric learning for image retrieval. This also explains why our method is focused mainly on image retrieval.

\begin{table}
\begin{floatrow}

\capbtabbox{
\scriptsize
\setlength{\tabcolsep}{0.7mm}
\begin{tabular}{|p{80 pt}|c|c|c|c|c|c|c|}
\hline
\multicolumn{1}{|c|}{Configurations} & \multicolumn{3}{c|}{Original classes} & \multicolumn{3}{c|}{New classes$ ^{\dagger}$} \\ \cline{1-7} 
\multicolumn{1}{|c|}{Recall@K(\%)}  & K=1  & K=2 & K=4 & K=1 & K=2 & K=4 \\ \hline
$\mathcal{A}$(1-100) (initial model) & 79.41 & 85.64 & 89.63 & - & - & -  \\
\hline 

+$\mathcal{B}$(101-125) w Our method  & 76.65 & 83.47 & 88.86 & 73.13 & 82.31 & 88.44 \\
\hline 
 +$\mathcal{B}$(101-200) w Our method & 74.41 & 82.57 & 88.52 & 73.11 & 80.84 & 86.64 \\
\hline  
$\mathcal{A}$(1-125) (reference model) &  77.84 & 83.94 & 87.80 & 79.25 & 85.54 & 91.96 \\     
\hline        
$\mathcal{A}$(1-200) (reference model) & 77.33 & 85.08 & 89.03 & 76.64 & 83.53 & 89.12 \\  
\hline  
\end{tabular}
}
{
\vspace{-2em}
 \caption{Recall@K (\%)  on the CUB-Birds dataset when 25 or 100 new classes are added at once. Correspondingly, $ ^{\dagger}$ indicates the results are tested on different new classes. }
 \label{Adding_number_cub_p1}
}

\capbtabbox{
\linespread{1.2}
\scriptsize
\setlength{\tabcolsep}{1.4mm}
\begin{tabular}{|c|c|c|c|c|c|}
\hline
\multicolumn{1}{|l|}{\multirow{2}{*}{Method}} & \multicolumn{5}{c|}{Number of new classes} \\ \cline{2-6} 
\multicolumn{1}{|l|}{}  & 20 & 40 & 60 & 80 & 100 \\ \hline
L2 loss  & 77.3   & 47.5  & 40.5  & 36.6  & 32.8  \\ \hline
EWC      & 77.3   & 60.5  & 50.9  & 43.3  & 39.5  \\ \hline
LwF      & 77.3   & 62.5  & 52.9  & 46.2  & 41.0  \\ \hline
Ours     & 77.3   & 64.6  & 55.8  & 49.2  & 43.3  \\ \hline
\end{tabular}
}
{
\vspace{-2em}
 \caption{ Average top-1 accuracy of incremental learning for image classification on CIFAR-100 dataset. }
 \label{results_for_classification}
}

\end{floatrow}
\end{table}

\subsection{Training Time Comparison}

 We compare the average training time on the CUB-Birds dataset when 100 new classes are added at once. The results are shown in Figure \ref{training_time_cub}. Note that all models in five methods are starting from the same initial model trained on the original 100 classes as initialization. The reference time is from joint training where the initial model is trained on all classes. The other four methods are incrementally learning the new classes only. We observe that our method saves more time by 50\% as expected. EWC and ALASSO algorithms take more time than reference because the gradients computation during back-propagation process is time-consuming.

\subsection{Components Analysis}
\noindent\textbf{Ablation Study.} We have done an ablation study on the CUB-Birds dataset when multiple classes are added at once. Note that the component ``$ L_{ce} +  L_{triplet} $'' comprises our baseline performance, thus we analyze the different loss items in Eq. \ref{Over_all_loss_functions}. We can observe the influence of difference components for the original and new classes. The results are shown in Table \ref{Ablation_study}.
\begin{table}[t] 
    \centering
    \scriptsize
    \setlength{\tabcolsep}{0.7mm}
    \begin{tabular}{|p{110 pt}|c|c|c|c|c|c|c|}
    \hline
    \multicolumn{1}{|c|}{Configurations} & \multicolumn{3}{c|}{Original classes} & \multicolumn{3}{c|}{New classes} \\ \cline{1-7} 
    \multicolumn{1}{|c|}{Recall@K(\%)}  & K=1  & K=2 & K=4 & K=1 & K=2 & K=4 \\ \hline
$\mathcal{A}$(1-100) (initial model)  &   79.41 & 85.64 & 89.63 & - & - & -  \\
\hline 

+$\mathcal{B}$(101-200) w $ L_{ce} +  L_{triplet} $  & 53.90 & 64.56 & 73.56 & 76.18 & 82.56 & 87.39 \\ 
+$\mathcal{B}$(101-200) w  $ L_{ce} +  L_{triplet} +  L_{dist} $  & 54.92 & 66.40 & 75.42 & 75.76 & 82.69 & 86.93 \\
+$\mathcal{B}$(101-200) w  $ L_{ce} +  L_{triplet} + L_{mmd} $  & 73.36 & 81.25 & 87.43 & 73.40 & 81.60 & 86.64 \\ 
+$\mathcal{B}$(101-200) w  $ \! L_{ce} \! + \! L_{triplet} \!\! + \!\!  L_{dist} \!\! + \!\! L_{mmd} $  & 74.41 & 82.57 & 88.52 & 73.11 & 80.84 & 86.64 \\ 
\hline         
$\mathcal{A}$(1-200) (reference model) & 77.33 & 85.08 & 89.03 & 76.64 & 83.53 & 89.12 \\     
\hline        
\end{tabular}
     \vspace{-2em}
    \caption{Ablation study for different components of loss function}
\label{Ablation_study} 
\end{table}

\noindent\textbf{Hyper-parameters Sensitivity Analysis.}
We explore the sensitivity of hyper-parameters $ \alpha, \beta $ in Eq. \ref{Over_all_loss_functions}, which affect significantly the trade-off performance. We conduct this experiment on the CUB-Birds dataset. As shown in Table \ref{Hyper_Parameter_analysis}, we find that the incrementally-trained model is more sensitive to $ \beta $ than $ \alpha $. For instance, when $ \alpha $ is set as 0.1, but $ \beta $ changes from 0.1 to 1, model $ \mathcal{B} $ performs better on the new classes and significantly retains its previous performance. However, this obvious trend cannot be observed when $ \beta $ is set as 0.1, but $ \alpha $ changes from 0.1 to 1 where the model $ \mathcal{B} $ performs almost the same on the original and new classes. Finally, if $ \alpha \!\! = \!\! \beta \!\! = \!\! 1 $, the incrementally-trained model $ \mathcal{B} $ keeps a better trade-off performance between the original and the new classes.
\begin{table}[!h] 
\scriptsize
\centering
\setlength{\tabcolsep}{0.7mm}
\begin{tabular}{|p{85 pt}|c|c|c|c|c|c|c|}
\hline
\multicolumn{1}{|c|}{Configurations} & \multicolumn{3}{c|}{Original classes} & \multicolumn{3}{c|}{New classes} \\ \cline{1-7} 
\multicolumn{1}{|c|}{Recall@K (\%)}  & K=1  & K=2 & K=4  & K=1 & K=2 & K=4 \\ \hline
$\mathcal{A}$(1-100) (initial model)   & 79.41  & 85.64 & 89.63 & - & - & -  \\
\hline 
+$\mathcal{B}$(101-200) ($\alpha \!\! = \!\! 0.1, \beta\!\! = \!\!0.1$)   & 56.53  & 66.31 & 75.59 & 77.52 & 83.82 & 88.15 \\
+$\mathcal{B}$(101-200) ($\alpha \!\! = \!\! 0.1, \beta\!\! = \!\!1$)   & 73.31 & 82.00 & 87.14 & 72.77 & 80.92 & 87.14  \\ 
+$\mathcal{B}$(101-200) ($\alpha \!\! = \!\! 0.1, \beta\!\! = \!\!10$)   &  79.58 & 85.76 & 90.47  & 49.50 & 61.51 & 70.59 \\            
+$\mathcal{B}$(101-200) ($\alpha \!\! = \!\! 1, \beta\!\! = \!\!0.1$)  & 55.81 & 67.25 & 75.59 & 77.02 & 83.91 & 87.90  \\   
+$\mathcal{B}$(101-200) ($\alpha \!\! = \!\! 1, \beta\!\! = \!\!1$)    & \underline{74.41} & \underline{82.57} & \underline{88.52}  & \underline{73.11} & \underline{80.84} & \underline{86.64} \\ 
+$\mathcal{B}$(101-200) ($\alpha \!\! = \!\! 1, \beta\!\! = \!\! 10 $)     & 79.41 & 86.31 & 90.51  & 48.82 & 61.09 & 71.05  \\ 
\hline         
$\mathcal{A}$(1-200) (reference model)  & 77.33 & 85.08 & 89.03  & 76.64 & 83.53 & 89.12 \\     
\hline        
\end{tabular}
    \vspace{-2em}
    \caption{ Sensitivity analysis of the hyper-parameters $ \alpha, \beta $. The better trade-off performance of the hyper-parameters are underlined. }
\label{Hyper_Parameter_analysis} 
\end{table}

\section{Conclusion}

 In this paper, for the first time, we have exploited incremental learning for fine-grained image retrieval in several scenarios for increasing numbers of image categories when only images of new classes are used. To overcome the catastrophic forgetting, we adopted the distillation loss function to constrain the classifier in the original network and the incremental classifier in the adaptive network. Moreover, we introduced a regularization function, based on Maximum Mean Discrepancy (MMD), to minimize the discrepancy between features of newly added classes from the original and the adaptive network. Comprehensive and empirical experiments on two fine-grained datasets show the effectiveness of our method that is superior over existing methods. In the future, it is promising to investigate incremental learning between different fine-grained datasets for image retrieval.

\section{Acknowledgment}

This work is supported by LIACS MediaLab at Leiden University, China Scholarship Council (CSC No. 201703170183), and the FWO project ``Structure from Semantic'' (grant No. G086617N). We would like to thank NVIDIA for the donation of GPU cards.

\bibliography{egbib}

\newpage

\LARGE \textbf{Supplementary Material}
\vspace{0.5em}
\normalsize

In this supplementary material, we provide additional quantitiative and qualitative results, which are not shown in the main paper. We tested on another dataset: Stanford-Dogs \cite{khosla2011novel}. Note that recorded results are under the same configurations with the CUB-Birds dataset.

\section*{1 One-step Incremental Learning for FGIR}

(1) Recall@K Evaluation on the Stanford-Dogs Dataset

The process includes two stages. First, we use the cross-entropy and triplet loss to train the network $ \mathcal{A} $ on the original classes (1-60), denoted as $ \mathcal{A}(1 \verb|-| 60)$. Second, only images of new classes are added at once to train network $ \mathcal{B}$, denoted as $ \mathcal{B}(61 \verb|-| 120)$. We observe similar trends as the results we shown in main paper, when our method achieves good performance on the original classes and new classes with Recall@1= 76.67\% and Recall@1=81.88\%, respectively. Compared to the initial model on the original classes, our method has dropped Recall@1 performance by 4.00\% (80.67\% $ \!\! \to \!\! $ 76.67\%). 


\begin{table}[!h] 
\centering
\scriptsize
\setlength{\tabcolsep}{0.9mm}
\begin{tabular}{|p{110 pt}|c|c|c|c|c|c|c|}
\hline
\multicolumn{1}{|c|}{Configurations} & \multicolumn{3}{c|}{Original classes} & \multicolumn{3}{c|}{New classes} \\ \cline{1-7} 
\multicolumn{1}{|c|}{Recall@K(\%)}  & K=1  & K=2 & K=4 & K=1 & K=2 & K=4 \\ \hline
$\mathcal{A}$(1-60) (initial model)     &   80.67 & 87.27 & 92.20 & - & - & -  \\
\hline 
+$\mathcal{B}$(61-120) w feature extraction   & - & - & - & 75.64 & 83.91 & 90.48 \\
+$\mathcal{B}$(61-120) w fine-tuning    & 61.43 & 72.80 & 81.70 & 78.93 & 86.99 & 91.55 \\
+$\mathcal{B}$(61-120) w LwF ($ L_{dist} $)    & 61.77 & 72.72 & 81.70 & 78.52 & 86.38 & 91.12 \\
+$\mathcal{B}$(61-120) w EWC    & 62.24 & 73.30 & 82.82 & 78.90 & 86.59 & 91.19 \\
+$\mathcal{B}$(61-120) w ALASSO    & 62.61 & 74.49 & 82.98 & 78.14 & 85.98 & 91.02 \\
+$\mathcal{B}$(61-120) w L2 loss   & 72.07 & 81.44 & 87.47 & \textbf{82.21} & 88.75 & 92.52 \\
+$\mathcal{B}$(61-120) w Our method  & \textbf{76.67} & \textbf{85.10} & \textbf{91.14} & 81.88 & \textbf{88.98} & \textbf{93.36} \\
\hline         
$\mathcal{A}$(1-120) (reference model) & 79.29 & 86.86 & 91.61 & 82.57 & 88.75 & 93.13 \\     
\hline            
\end{tabular}
    \caption{ Recall@K (\%) of incremental FGIR on the Stanford-Dogs dataset when new classes are added at once. The best performance are in bold. }
\label{Results_Table2}
\end{table}

 (2) Precision-Recall Curves and mAP Results
 
  We report the precision-recall curves and mAP results in Figure \ref{Precision_curve_all}. We can observe these curves share with the similar trends with those from the CUB-Birds dataset.  Overall, our method can effectively address the catastrophic forgetting issue on the original classes while achieve ideal performance on the new classes.
  

\begin{figure}[!h]
\centering  
  \subfigure[]
 { \label{without_adversary}     
   \includegraphics[width=0.225\columnwidth]{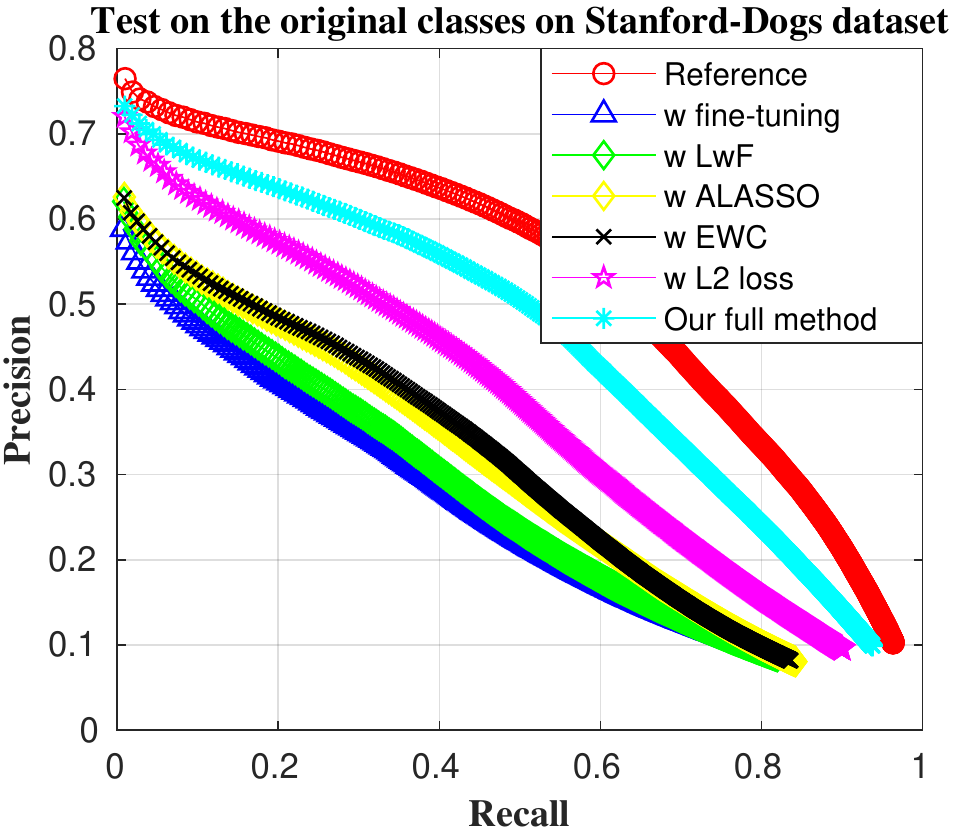}  
 }     
  \subfigure[] 
 { \label{Full_method_choosed}   
   \includegraphics[width=0.22\columnwidth]{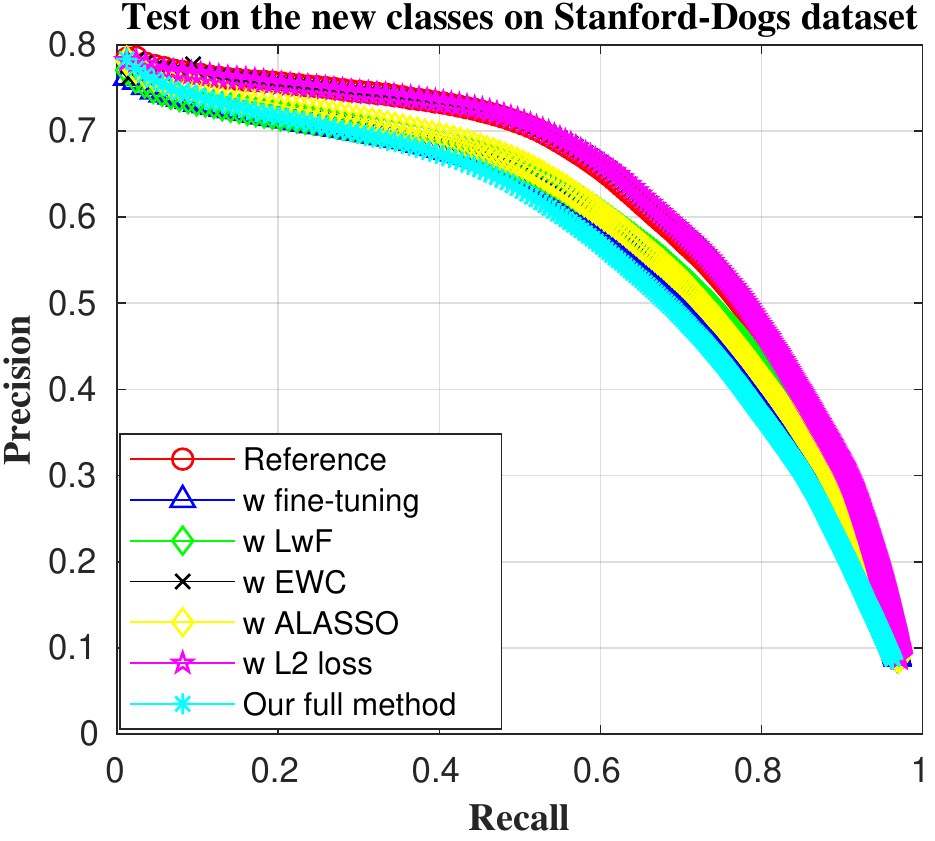}   
  }
    \subfigure[] 
 { \label{Without_KL}   
   \includegraphics[width=0.23\columnwidth]{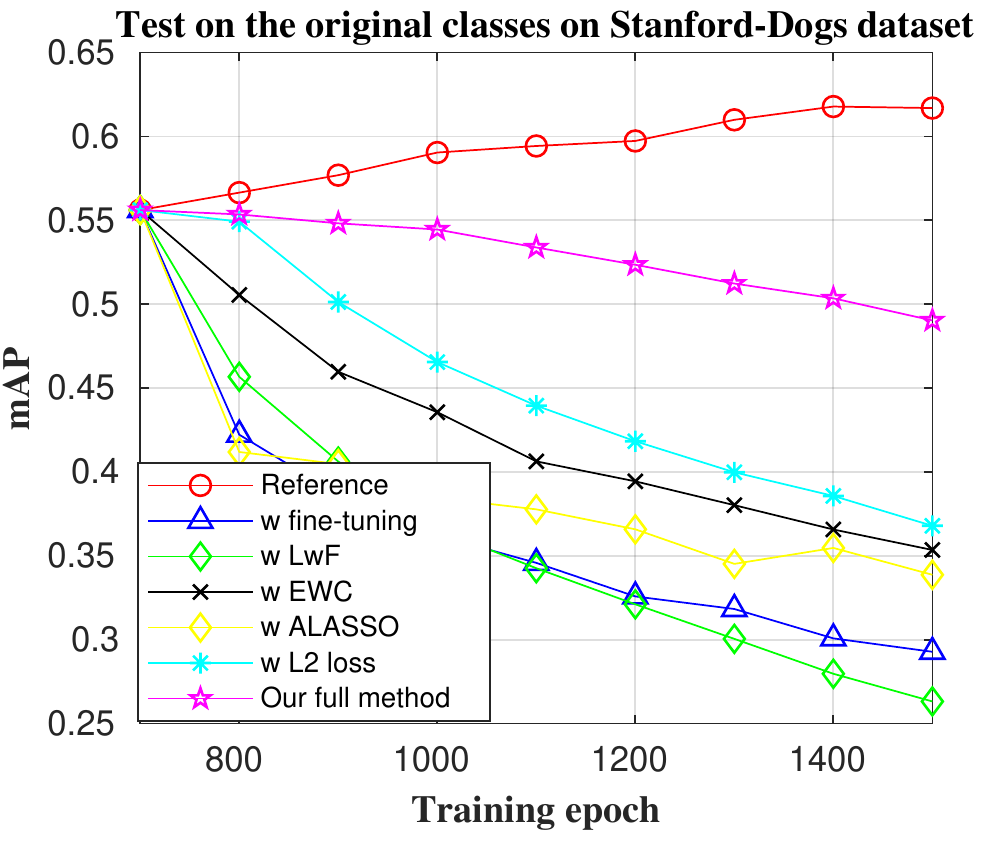}   
  }  
  \subfigure[] 
 { \label{training_time_dog}     
   \includegraphics[width=0.23\columnwidth]{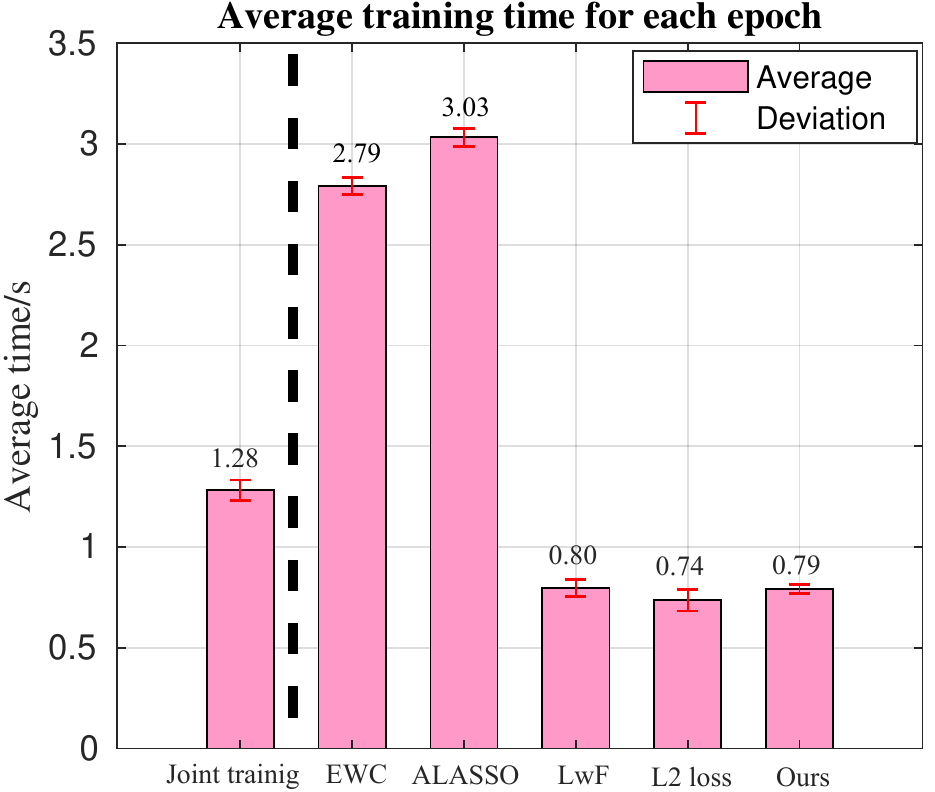} 
 }  

     \caption{ Figure (a)-(b) denote the precision-recall curves tested on the original classes and new classes on the Stanford-Dogs dataset. The larger the area under each curve, the better performance of the method. Figure (c) depicts the mAP results for different methods as the training proceeds. We only show the results tested on the original classes. Being closer to the reference curve (red one) indicates less performance degradation, \emph{i.e.}, the method can maintain its previous performance on the original classes on the Stanford-Dogs dataset.}  
      \vspace{-1.5em}
\label{Precision_curve_all}  
 \end{figure}

 (3) t-SNE Visualization for Feature Distribution
 
\begin{figure}[!h]

\centering  
  \subfigure[]
 { \label{Baseline1}     
   \includegraphics[width=0.45\columnwidth]{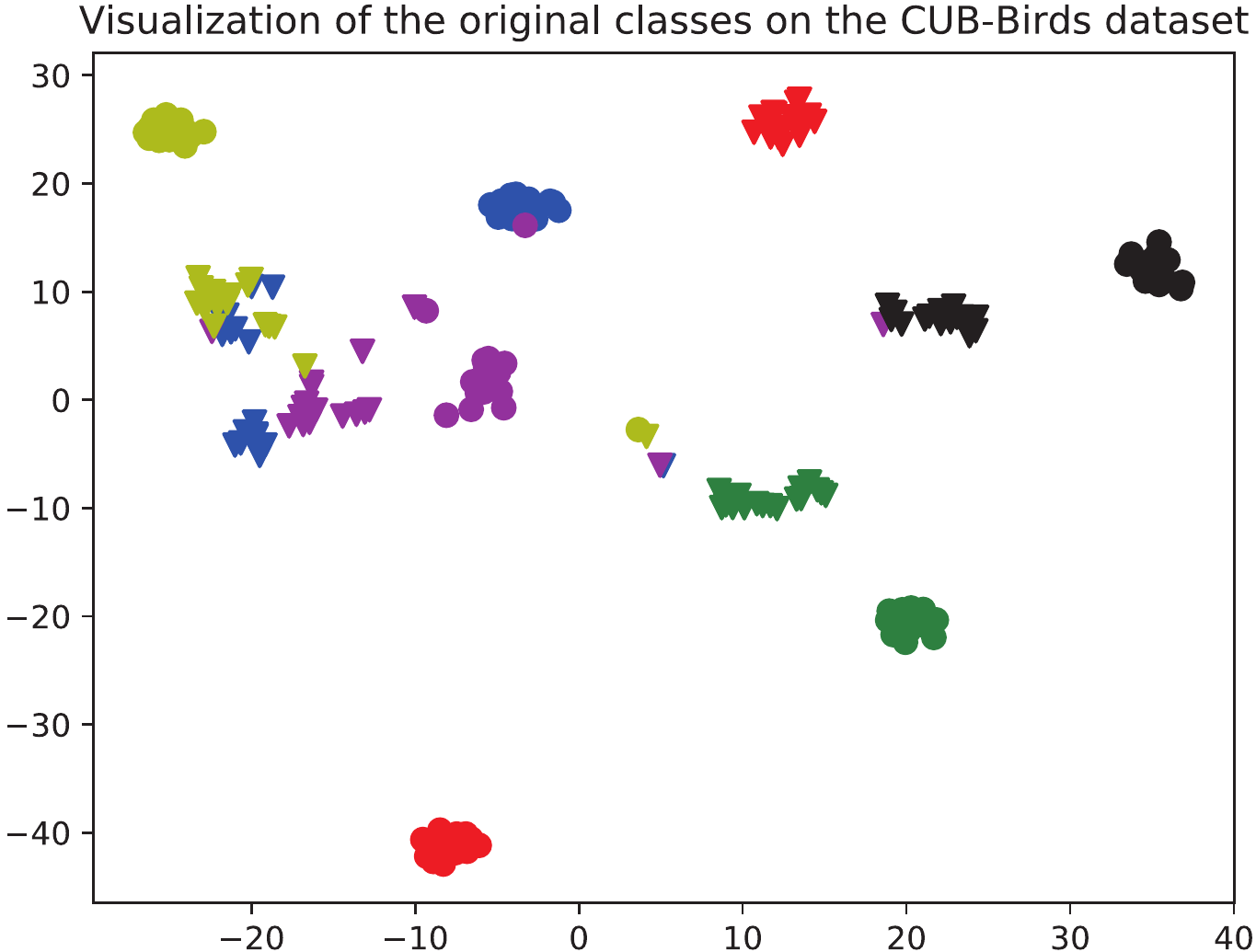}  
 }     
  \subfigure[] 
 { \label{Without_KL}   
   \includegraphics[width=0.45\columnwidth]{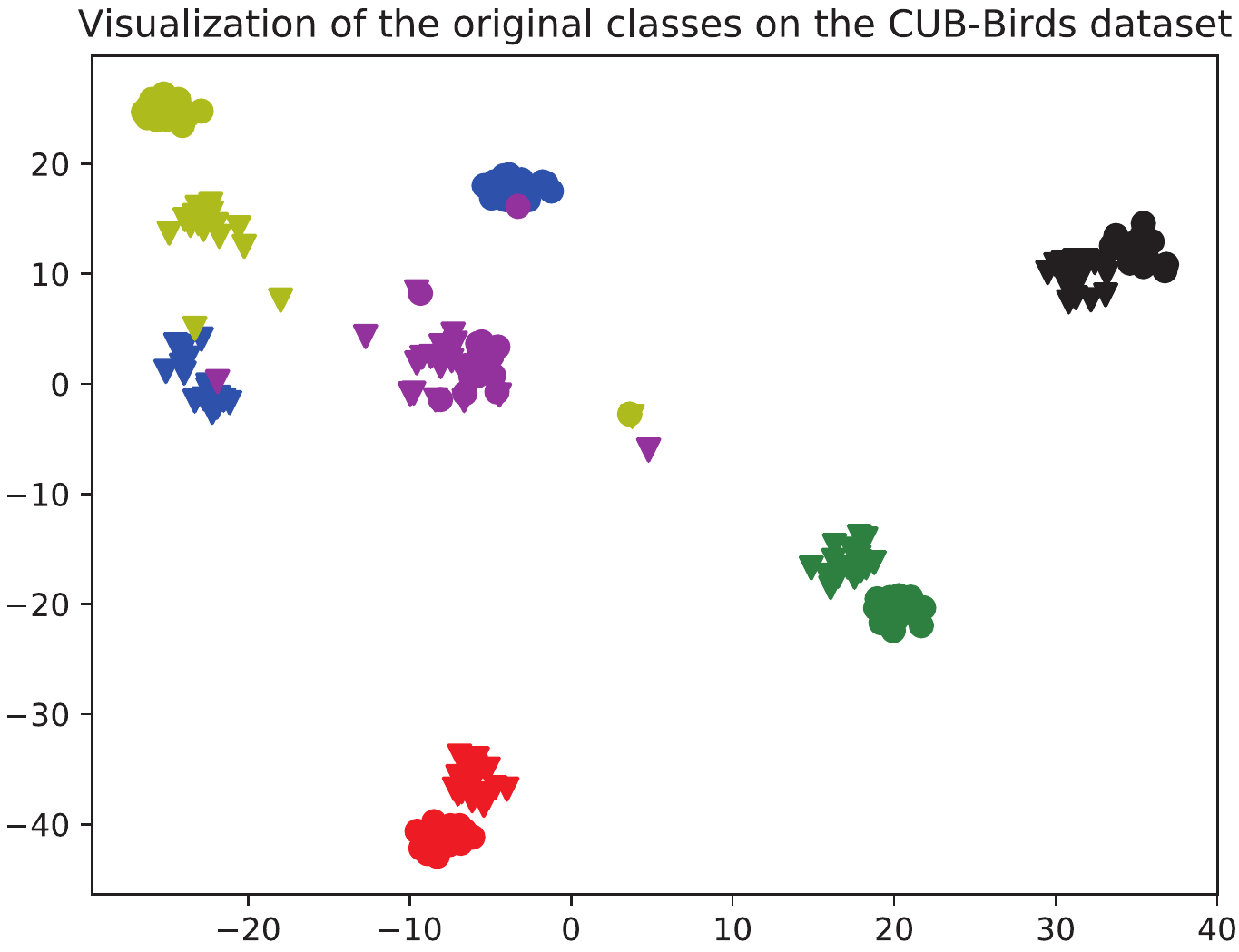}   
  }  

  \caption{ t-SNE visualization for feature distribution of 6 categories. The circle shape indicates the features from reference model, which has the same distribution in two cases. The triangle shape denotes the feature from models trained with/without MMD loss.  (a): model trained without MMD loss; (b): model trained with MMD loss.  }  
\label{t_SNE_visualization}   
 \end{figure}

We visualize the feature distributions with and without MMD loss in Figure \ref{t_SNE_visualization}, which demonstrate the MMD loss reduces the distance between distributions and effectiveness for mitigating the forgetting issue.

 \section*{2 Influence of Added Multiple Classes}

 In previous experiments, we add multiple classes (\emph{i.e.} 100 new classes for the CUB-Birds dataset) for one-step incremental training at once. Herein, we further explore the influence of the new classes number for the Stanford-Dogs dataset where we choose 60 new classes and 5 classes for incremental learning.

The results are reported in Table \ref{Adding_number_dog}. We observe these two datasets share with similar trends that larger new coming classes lead to heavier forgetting issue. For the Stanford-Dogs dataset, when only 5 new classes are added, the Recall@1 drops from 80.67\% to 79.75\%, compared to the one drops from 80.67\% to 76.67\%
when 60 new classes are added.
 

\begin{table}[!t]
\centering
\scriptsize
\setlength{\tabcolsep}{0.7mm}
\begin{tabular}{|p{100 pt}|c|c|c|c|c|c|c|}
\hline
\multicolumn{1}{|c|}{Configurations} & \multicolumn{3}{c|}{Original classes} & \multicolumn{3}{c|}{New classes$ ^{\dagger}$} \\ \cline{1-7} 
\multicolumn{1}{|c|}{Recall@K(\%)}  & K=1  & K=2 & K=4 & K=1 & K=2 & K=4 \\ \hline
$\mathcal{A}$(1-60) (initial model) & 80.67 & 87.27 & 92.20 & - & - & -  \\
\hline 
\quad  +$\mathcal{B}$(61-65) w Our full method  & 79.75 & 87.23 & 91.92 & 97.45 & 98.55 & 99.27 \\
\hline  
\quad +$\mathcal{B}$(61-120) w Our full method & 76.67 & 85.10 & 91.14 & 81.88 & 88.98 & 93.36 \\
\hline
$\mathcal{A}$(1-65) (reference model) & 79.62  & 86.15 & 90.91 & 96.73 & 97.82 & 98.55  \\ 
\hline         
$\mathcal{A}$(1-120) (reference model) & 79.29 & 86.86 & 91.61 & 82.57 & 88.75 & 93.13 \\   
\hline            
\end{tabular}
    \caption{ Recall@K (\%)  on the Stanford-Dogs dataset when 5 or 60 new classes are added at once. Correspondingly, $ ^{\dagger}$ indicates the results are tested on different new classes. }
\label{Adding_number_dog}
\end{table}

 \section*{3 Multi-step Incremental Learning for FGIR}
 
  We report the results on the Stanford-Dogs dataset in Table \ref{Results_Table4} when new classes are added sequentially. We observe similar trends as those for the CUB-Birds dataset. Compared to the other two methods, the proposed method has ideal retrieval performance on the newly added classes and the original classes. 
 
\begin{table*}[!h]
\tiny
\centering
\setlength{\tabcolsep}{0.75mm}
\begin{tabular}{|p{30pt}|l|c|c|c|c|c|c|c|c|c|c|c|c|c|c|c|}
\hline
\multicolumn{2}{|c|}{Configurations} & \multicolumn{3}{c|}{Original (1-60)} & \multicolumn{3}{c|}{Added new (61-75)} & \multicolumn{3}{c|}{Added new (76-90)} & \multicolumn{3}{c|}{Added new (91-105)} & \multicolumn{3}{c|}{Added new (106-120)}\\ \cline{1-17} 

\multicolumn{2}{|c|}{Recall@K(\%)} & \multicolumn{1}{c|}{K=1} & \multicolumn{1}{c|}{K=2} & \multicolumn{1}{c|}{K=4} & \multicolumn{1}{c|}{K=1} & \multicolumn{1}{c|}{K=2} & \multicolumn{1}{c|}{K=4}& \multicolumn{1}{c|}{K=1} & \multicolumn{1}{c|}{K=2} & \multicolumn{1}{c|}{K=4} & \multicolumn{1}{c|}{K=1} & \multicolumn{1}{c|}{K=2} & \multicolumn{1}{c|}{K=4} & \multicolumn{1}{c|}{K=1} & \multicolumn{1}{c|}{K=2} & \multicolumn{1}{c|}{K=4} \\ \hline

\multicolumn{2}{|c|}{$\mathcal{A}$(1-60) (initial model)} & \multicolumn{1}{c|}{80.67} & \multicolumn{1}{c|}{87.27} & \multicolumn{1}{c|}{92.20}& \multicolumn{1}{c|}{-} & \multicolumn{1}{c|}{-} & \multicolumn{1}{c|}{-}  & \multicolumn{1}{c|}{-}  & \multicolumn{1}{c|}{-}  & \multicolumn{1}{c|}{-} & \multicolumn{1}{c|}{-}  & \multicolumn{1}{c|}{-}  & \multicolumn{1}{c|}{-}  & \multicolumn{1}{c|}{-}  & \multicolumn{1}{c|}{-}  & \multicolumn{1}{c|}{-}       \\ \hline

\multirow{5}{*}{\begin{tabular}[c]{@{}c@{}}LwF\\ algorithm \cite{li2017learning} \end{tabular}} & \multicolumn{1}{l|}{+$\mathcal{B}$(61-75)} & \multicolumn{1}{c|}{50.87} & \multicolumn{1}{c|}{62.76} & \multicolumn{1}{c|}{73.40}    & \multicolumn{1}{c|}{88.35} & \multicolumn{1}{c|}{92.48} & \multicolumn{1}{c|}{94.36} & \multicolumn{1}{c|}{-} & \multicolumn{1}{c|}{-} & \multicolumn{1}{c|}{-} & \multicolumn{1}{c|}{-} & \multicolumn{1}{c|}{-} & \multicolumn{1}{c|}{-} & \multicolumn{1}{c|}{-} & \multicolumn{1}{c|}{-}  & \multicolumn{1}{c|}{-}\\

 &  +$\mathcal{B}$(61-75)(76-90) & 42.06 & 53.62 & 65.10 & 71.18 & 82.46 & 88.10 & 77.33 & 87.19 & 92.44  & - & - & - & - & - & -  \\
 &  +$\mathcal{B}$(61-75)(76-90)(91-105) &   37.58 & 50.48 & 63.00 & 60.65 & 73.68 & 82.33 & 70.10 & 81.38 & 87.40  & 80.62 & 87.17 & 92.48 & - & - & -  \\
 &  +$\mathcal{B}$(61-75)(76-90)(91-105)(106-120) &  38.46  & 50.63 & 62.59 & 59.90 & 72.68 & 81.20 & 63.86 & 77.22 & 85.54  & 68.41 & 77.70 & 85.66 & 81.34 & 88.69 & 92.74  \\
 \cline{2-17}
  \hline
 
\multirow{5}{*}{\begin{tabular}[c]{@{}c@{}}EWC\\  algorithm \cite{kirkpatrick2017overcoming} \end{tabular}} & \multicolumn{1}{l|}{+$\mathcal{B}$(61-75)} & \multicolumn{1}{c|}{55.84} & \multicolumn{1}{c|}{67.64} & \multicolumn{1}{c|}{77.57}    & \multicolumn{1}{c|}{89.10} & \multicolumn{1}{c|}{92.61} & \multicolumn{1}{c|}{94.36} & \multicolumn{1}{c|}{-} & \multicolumn{1}{c|}{-} & \multicolumn{1}{c|}{-} & \multicolumn{1}{c|}{-} & \multicolumn{1}{c|}{-} & \multicolumn{1}{c|}{-} & \multicolumn{1}{c|}{-} & \multicolumn{1}{c|}{-}  & \multicolumn{1}{c|}{-}\\

 &  +$\mathcal{B}$(61-75)(76-90) & 45.32 & 58.29 & 68.85 & 79.82 & 85.21 & 90.35 & 81.38 & 88.72 & 93.32  & - & - & - & - & - & -  \\
 &  +$\mathcal{B}$(61-75)(76-90)(91-105) & 37.60 & 49.71 & 61.88 & 67.04 & 79.04 & 85.71 & 67.47 & 79.52 & 88.39  & 81.33 & 86.99 & 91.15 & - & - & -  \\
 &  +$\mathcal{B}$(61-75)(76-90)(91-105)(106-120) &  34.08  & 45.60 & 58.40 & 63.53 & 75.19 & 83.71 & 63.42 & 77.66 & 86.31  & 70.00 & 79.12 & 85.84 & 81.99 & 87.96 & 92.10  \\
 \cline{2-17} 
  \hline
  
\multirow{5}{*}{\begin{tabular}[c]{@{}c@{}}L2 loss\\  algorithm  \cite{michieli2019incremental} \end{tabular}} & \multicolumn{1}{l|}{+$\mathcal{B}$(61-75)} & \multicolumn{1}{c|}{65.30} & \multicolumn{1}{c|}{75.83} & \multicolumn{1}{c|}{83.51}    & \multicolumn{1}{c|}{90.85} & \multicolumn{1}{c|}{94.74} & \multicolumn{1}{c|}{95.61} & \multicolumn{1}{c|}{-} & \multicolumn{1}{c|}{-} & \multicolumn{1}{c|}{-} & \multicolumn{1}{c|}{-} & \multicolumn{1}{c|}{-} & \multicolumn{1}{c|}{-} & \multicolumn{1}{c|}{-} & \multicolumn{1}{c|}{-}  & \multicolumn{1}{c|}{-}\\

 &  +$\mathcal{B}$(61-75)(76-90) & 55.97 & 67.36 & 77.04 & 84.46 & 90.73 & 92.73 & 80.94 & 89.38 & 93.54  & - & - & - & - & - & -  \\
 &  +$\mathcal{B}$(61-75)(76-90)(91-105) & 50.38 & 62.87 & 73.64 & 72.68 & 82.21 & 88.85 & 75.68 & 84.67 & 91.57  & 83.72 & 90.00 & 93.81 & - & - & -  \\
 &  +$\mathcal{B}$(61-75)(76-90)(91-105)(106-120) &  46.01  & 58.74 & 69.64 & 67.79 & 78.07 & 85.71 & 72.51 & 84.45 & 90.03  & 74.87 & 83.98 & 89.82 & 86.21 & 91.36 & 94.39  \\
 \cline{2-17} 
 
  \hline
  
\multirow{4}{*}{\begin{tabular}[c]{@{}c@{}}Our method \end{tabular}} & \multicolumn{1}{l|}{+$\mathcal{B}$(61-75)} & \multicolumn{1}{c|}{76.07} & \multicolumn{1}{c|}{84.88} & \multicolumn{1}{c|}{90.11} & \multicolumn{1}{c|}{91.85} & \multicolumn{1}{c|}{95.36} & \multicolumn{1}{c|}{96.87} & \multicolumn{1}{c|}{-} & \multicolumn{1}{c|}{-} & \multicolumn{1}{c|}{-} & \multicolumn{1}{c|}{-} & \multicolumn{1}{c|}{-} & \multicolumn{1}{c|}{-} & \multicolumn{1}{c|}{-} & \multicolumn{1}{c|}{-} & \multicolumn{1}{c|}{-}\\

 &  +$\mathcal{B}$(61-75)(76-90) & 70.67 & 80.48 & 87.87 & 89.10 & 93.11 & 95.99 & 84.23 & 89.92 & 93.43  & - & - & - & - & - & -  \\
 &  +$\mathcal{B}$(61-75)(76-90)(91-105) & 67.75 & 79.17 & 86.45 & 86.09 & 91.98 & 95.49 & 81.60 & 90.25 & 93.76  & 84.25 & 89.03 & 93.45 & - & - & -  \\
 &  +$\mathcal{B}$(61-75)(76-90)(91-105)(106-120) & 65.47  & 76.52 & 85.08 & 83.21 & 89.35 & 93.73 & 79.19 & 87.84 & 93.32  & 82.83 & 89.20 & 94.42 & 87.13 & 92.10 & 94.39  \\
 \cline{2-17}
  \hline
  
\multicolumn{2}{|c|}{$\mathcal{A}$(1-120) (reference model)}  &  79.29  & 86.86  &  91.61 & 92.61  & 94.99  & 96.37  &   82.48 & 90.80  &  93.76 & 83.72  & 91.33 &  95.58 & 86.12  &  93.11  & 95.96                 
\\ \hline
\end{tabular}

    \caption{ Recall@K (\%) results on the Stanford-Dogs dataset when new classes are added sequentially. ``Added new (61-75)'' indicates we use first 15 classes (61-75) as the first incremental part to train the network.}
\label{Results_Table4}
\end{table*}

\end{document}